\documentclass[10pt, conference, letterpaper]{IEEEtran}
\IEEEoverridecommandlockouts

\usepackage[letterpaper]{geometry}
\newgeometry{left=1.57cm,right=1.57cm,bottom=2.4cm,top=2.0cm}
\usepackage{setspace}
\usepackage{booktabs} 
\usepackage{amsmath}
\usepackage{amssymb}
\usepackage{amsfonts}
\usepackage{float}
\usepackage{graphicx}
\usepackage{xcolor}
\usepackage{colortbl}
\usepackage{multirow}
\usepackage{mathtools}
\usepackage{algorithm}
\usepackage{algorithmicx}
\usepackage{algpseudocode}
\usepackage{tabto}
\usepackage{todonotes}
\usepackage{hhline}
\usepackage{enumitem}
\usepackage[belowskip=-8pt,aboveskip=4pt,font=footnotesize,labelfont={it}]{caption}
\usepackage{subfigure}
\renewcommand{\baselinestretch}{.97}

\usepackage{tikz}

\begin{document}
\title{FAdeML: Understanding the Impact of Pre-Processing Noise Filtering on Adversarial Machine Learning}
\author{\IEEEauthorblockN{Faiq Khalid\IEEEauthorrefmark{1}, 
        Muhammad Abdullah Hanif\IEEEauthorrefmark{1},
		Semeen Rehman\IEEEauthorrefmark{1},
        Junaid Qadir\IEEEauthorrefmark{2},
		Muhammad Shafique\IEEEauthorrefmark{1}}
 	\IEEEauthorblockA{\IEEEauthorrefmark{1}Vienna University of Technology, Vienna, Austria}
 	\IEEEauthorblockA{\IEEEauthorrefmark{2}Information Technology University, Lahore, Pakistan}
 	\IEEEauthorblockA{Email: \{faiq.khalid,muhammad.hanif,semeen.rehman, muhammad.shafique\}@tuwien.ac.at,  junaid.qadir@itu.edu.pk}
}

\maketitle
\begin{abstract}
	Deep neural networks (DNN)-based machine learning (ML) algorithms have recently emerged as the leading ML paradigm particularly for the task of classification due to their superior capability of learning efficiently from large datasets. The discovery of a number of well-known attacks such as dataset poisoning, adversarial examples, and network manipulation (through the addition of malicious nodes) has, however, put the spotlight squarely on the lack of security in DNN-based ML systems. In particular, malicious actors can use these well-known attacks to cause random/targeted misclassification, or cause a change in the prediction confidence, by only slightly but systematically manipulating the environmental parameters, inference data, or the data acquisition block. Most of the prior adversarial attacks have, however, not accounted for the pre-processing noise filters commonly integrated with the ML-inference module. Our contribution in this work is to show that this is a major omission since these noise filters can render ineffective the majority of the existing attacks, which rely essentially on introducing adversarial noise. Apart from this, we also extend the state of the art by proposing a novel pre-processing noise \textit{F}ilter-aware \textit{Ad}v\textit{e}rsarial \textit{ML} attack called \textit{FAdeML}. To demonstrate the effectiveness of the proposed methodology, we generate an adversarial attack image by exploiting the ``VGGNet'' DNN trained for the ``German Traffic Sign Recognition Benchmarks (GTSRB)'' dataset, which despite having no visual noise, can cause a classifier to misclassify even in the presence of pre-processing noise filters.
\end{abstract}

\section{Introduction}\label{Intro}

Machine learning (ML) has been the great success story of the last decade. In particular, deep neural networks (DNN), which can be efficiently trained to eke out the maximum information from ``big data'', is the standout ML framework that has revolutionized diverse fields such as object recognition, 
information retrieval, signal processing (including video, image, and speech processing), and autonomous systems (with self-driving cars a prominent example). The impressive performance of DNN-based ML algorithms can be gauged from the fact that DNNs now regularly outperform human beings in a rapidly increasing number of domains that were historically considered amenable only for human analysis and outside the reach of machine algorithms---e.g., NLP, emotion recognition, and games such as Go. \textit{But despite their great prowess and success, DNN-based ML algorithms suffer from a critical problem:} these algorithms, as they are currently designed, are  particularly vulnerable to security attacks from malicious adversaries \cite{papernot2016limitations}.

A major reason behind this security vulnerability of ML algorithms stems from their implicit assumption that the testing or inference data will be similar in distribution to the training data, and that the model output would be sought in good faith by a trusted benign interacting party. Unfortunately this assumption flies in the face of any adversarial attempts to compromise the ML system, where a mismatch between the distributions on which the model is being trained and test is purposefully sought. We can safely anticipate that incorporation of ML models in critical settings (such as transportation, power grids) 
will make them an immediate target for malevolent adversaries 
who will be motivated to compromise these ML models and inflict massive damage. Addressing the lack of security of ML algorithms, therefore, assumes paramount importance and requires immediate attention from the community.

DNNs in particular have been shown to be very vulnerable to adversarial attacks. A striking example of DNN's vulnerability to \textit{adversarial examples}\footnote{``Adversarial examples'' are minor perturbations of the input (so minor that the changes could be visually imperceptible) especially crafted by the adversary purposefully to maximize the prediction error \cite{szegedy2013intriguing,goodfellow2018making}.} was first brought to prominence in 2013/14 through the work of Szegedy et al. \cite{szegedy2013intriguing} who showed that deep networks used for object recognition can be easily fooled through input images that have been perturbed in an imperceptible manner.\footnote{Although the adversarial vulnerability of DNNs was pointed out in 2013/14, the broader work on adversarial pattern recognition goes back further, at least till 2004 as outlined by Roli et al. in \cite{biggio2018wild}} \textit{Adversarial ML} is now a burgeoning field attracting significant attention \cite{goodfellow2018making}. An elaborate coverage of the history of adversarial machine learning and the issues involved can be seen in recent book-level treatments of adversarial ML \cite{vorobeychik2018adversarial,joseph_nelson_rubinstein_tygar_2018}.

\subsection{Types of Security Attacks on Machine Learning}

With reference to Fig.~\ref{fig:ML_Sec}, the major types of ML security attacks are: 

\begin{figure}[!t]
	\centering
	\includegraphics[width=1\linewidth]{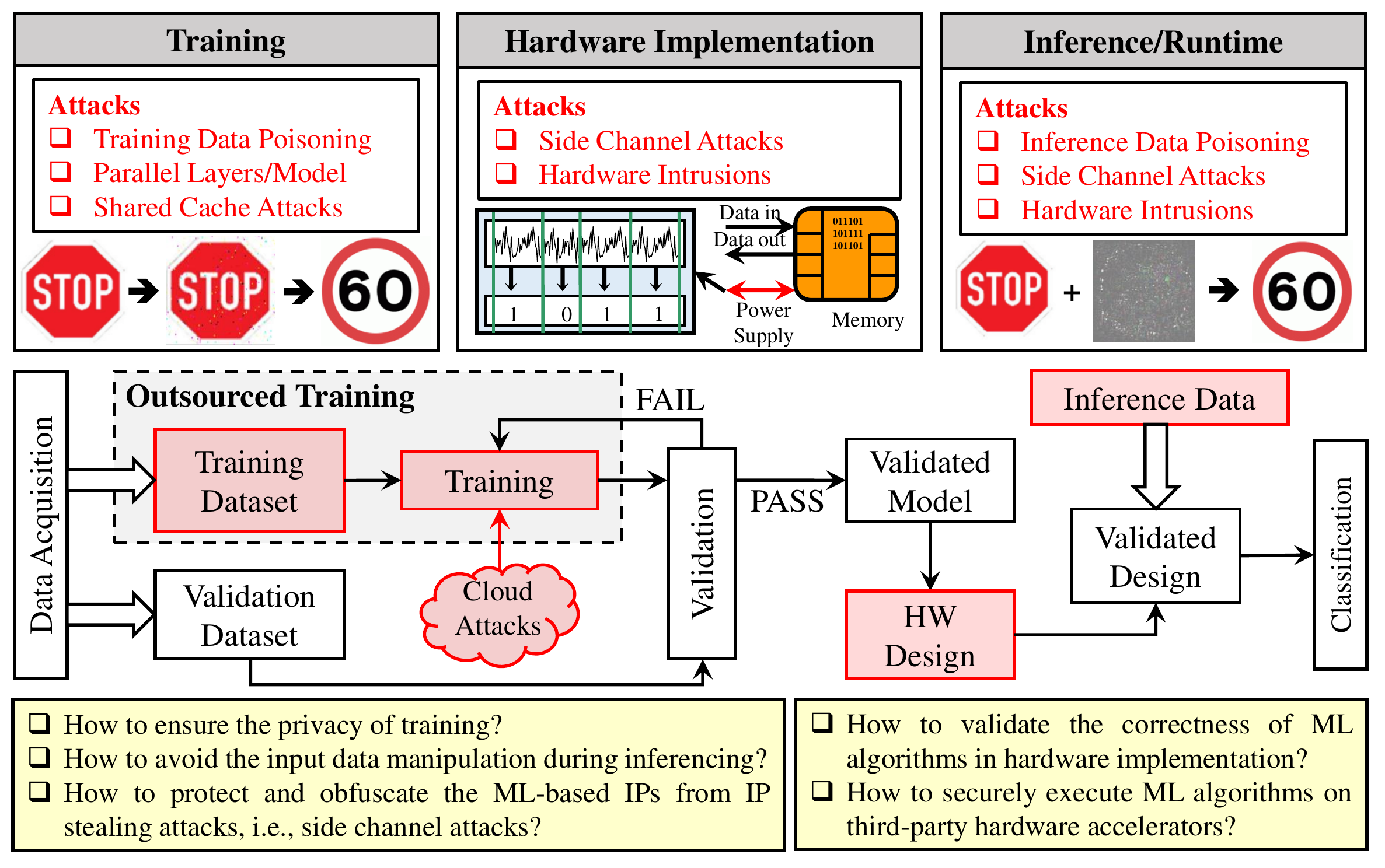} 
	\caption{\textit{An overview of security threats/attacks and their respective payloads for ML algorithms during training and inference \cite{shafique2018overview}}}
	\label{fig:ML_Sec}
\end{figure}

    
\begin{figure*}[!t]
	\centering
	\includegraphics[width=1\linewidth]{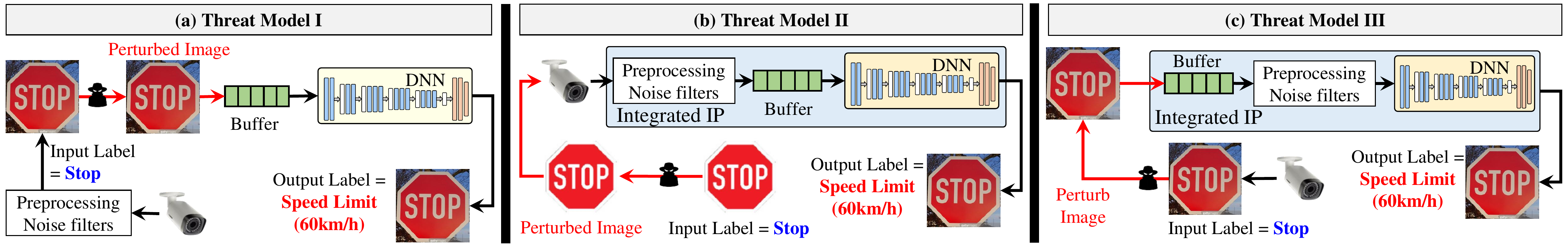}
	\caption{\textit{Threat Models: different attack scenarios based on the attack's methodology and access to the different blocks during the ML inference}}
	\label{fig:TM}
\end{figure*}

\vspace{1mm}    
\textit{Firstly}, \textbf{training attacks} or \textbf{poisoning attacks}, an attacker can adversarially perturb (or `poison') the training dataset, or otherwise manipulate the learning model/architecture or tool, with the goal of maximizing the classification error and denying service (e.g., by classifying a benign user as an intruder) through the injection of samples into the training data. 
For example, an attack can modify the underlying network structure through addition or deletion of parallel layers or neurons. In addition, when the training is being outsourced to some remote provider, other attacks (such as remote side-channel) can be used to steal the intellectual property (IP). 
    
\vspace{1mm}    
\textit{Secondly}, \textbf{inference attacks} or \textbf{evasion attacks}, the attacker can attempt to steal the IP through a side channel or through successive polling. The attacker can also perturb the inference data to create ``adversarial examples''. The goal in such attacks is evasion at the test time or the inference time through the manipulation of test samples \cite{biggio2013evasion,papernot2018sok,shafique2018overview}. 
Other possible attackers can be indirect beneficiaries who can either manipulate the inference data or intrude the hardware. Moreover, attackers can also perform side-channel attacks for IP stealing (see Fig. \ref{fig:ML_Sec}). In addition, the use of adversarial examples is an importance case of an inference stage attack \cite{xu2017feature,papernot2016cleverhans,brendel2017decision}. 
    
\vspace{1mm}    
\textit{Lastly}, \textbf{hardware implementation attacks}, the attacker seeks to exploit the hardware security threats, such as the manipulation of the hardware implementation of the trained ML model (hardware Trojans, \cite{zou2018potrojan}) and IP stealing (e.g., through side-channel or remote cyber-attacks) \cite{papernot2017practical,shokri2017membership}. Although a hardware intrusion can be used to launch several attack payloads, it requires complete access to the hardware blocks, without which these attacks are difficult or impossible to perform. Moreover, several sophisticated defense mechanism against the hardware attacks are becoming available.
    







\subsection{Challenges in Resisting ML Security Attacks}

Unlike the traditional systems, the security of ML-based systems is data dependent, which makes the job of ensuring security more challenging. One of the most common attacks is to exploit the data-dependency by manipulating/intruding the training dataset \cite{zhao2018data,wang2018data,jagielski2018manipulating} or the corresponding labels \cite{shafahi2018poison}. Similarly, the baseline ML algorithm and training tools can also be attacked by adding new layers/nodes or manipulating the hyper-parameters \cite{zou2018potrojan,li2018hu,liu2018sin,liao2018server}, as illustrated in Fig. \ref{fig:ML_Sec}. Moreover, in the case of the outsourced training, remote side-channel or cyber-attacks can be used to steal the IPs. Although data poisoning and ML algorithms/tools/architectures manipulation of attacks are quite effective, their effectiveness is limited by their substantial assumption of having complete access to information about the underlying ML architectures/baseline-model and the training dataset. 

\subsection{Motivating Pre-processing Noise Filter-Aware Adversarial ML}

Although most current adversarial ML security attacks incorporate pre-processing elements (such as shuffling, gray scaling, local histogram utilization and normalization) \cite{rakin2018robust} in their design and assume that an attacker can access the output of the pre-processing noise filtering, getting this access requires hardware manipulations and is practically difficult. If an attacker, on the other hand, does not have hardware access to the pre-processing filters, it becomes very challenging to incorporate the effect of pre-processing along with noise filtering, which raises the following key research questions:

\begin{enumerate}
    \item How can we analyze the impact of pre-processing noise filtering on adversarial examples?
    \item How can we incorporate the effects of the pre-processing noise filtering effects in the design of an improved adversarial ML attack?
\end{enumerate}


\subsection{Novel Contributions}
 
 The major contributions of this paper are:

	
\vspace{1mm}	
\textit{Firstly}, we provide an \textbf{elaborate analysis} on the impact of the pre-processing noise filtering on existing adversarial ML attack strategies (Section \ref{fademl}). We demonstrate that most state-of-the-art adversarial ML attacks on classification can be neutralized using pre-processing noise filters. Based on this analysis, and our anticipation that future attackers will evolve new strategies to defeat or even leverage pre-processing noise filters, we have made our next contribution.

\vspace{1mm}
\textit{Secondly}, we propose a \textbf{new attack methodology}, called pre-processing noise-\textit{F}ilter-aware \textit{Ad}v\textit{e}rsarial ML or simply \textit{FAdeML}, which is able to exploit pre-processing noise filtering as part of its attack strategy (Section \ref{FAde_attack}). \textit{As far as we know, this is the first work that has explicitly exploited noise filtering as part of an attack strategy on the ML security}. We demonstrate the effectiveness of the proposed FAdeML attack methodology by analyzying state-of-the-art adversarial attacks on the ``VGGNet'' \cite{wang2015places205} DNN trained for the ``German Traffic Sign Recognition Benchmarks (GTSRB)'' dataset \cite{stallkamp2011german} and show that FAdeML can force a classifier to misclassify even in the presence of pre-processing noise filters without any perceptible visual noise or change in the overall accuracy of the DNN.

\textbf{Open-source contributions:} We will release our design files, models, and attacks sources online at: \textit{http://LinkHiddenForBlindReview} for reproducible research.
	


\section{Background: Threat Models and Adversarial Attacks}\label{impact}


In this section, we will introduce the various attack/threat models and then some common state-of-the-art adversarial ML attacks.

\subsection{Attack Threat Models} 
\label{sec:AttackThreatModel}

In order to systematically reason about security, we need to articulate the threat model we are assuming: \textit{who are the possible attackers? what is the intention of the attack? what are the potential attack mechanisms? is the attack targeted or indiscriminate?} \cite{barreno2006can}. It is usually a good practice to adopt a conservative security attack model \cite{joseph_nelson_rubinstein_tygar_2018} and access for the ``worst-case’’ scenarios, but the best strategy is to assess the system security using different assumptions about the level of the adversary’s capability. In this paper, we assume the following three threat models:

\begin{enumerate}
    
    \vspace{1mm}
    \item \textbf{Threat Model I:} An attacker \textit{has access to the output of the pre-processing noise filter} and can perturb the image before storing it into the input buffer of the ML modules; see Fig. \ref{fig:TM}(a).

    \vspace{1mm}
    \item \textbf{Threat Model II:} An attacker \textit{does not have access to the output of the pre-processing noise filter or input buffers of the ML module} but it can manipulate the data before acquisition; see Fig. \ref{fig:TM}(b).
    
    \vspace{1mm}
    \item \textbf{Threat Model III:} An attacker \textit{does not have access to the pre-processing noise filter but it can directly perturb the acquired data} before storing it into the input buffer of the ML modules; see Fig. \ref{fig:TM}(c).

\end{enumerate}

\subsection{State-of-the-Art Adversarial ML Attacks}

An adversarial ML attack typically attempts to either reduce prediction confidence or to cause a (random or targeted) misclassification 
by adding an imperceptible purposefully crafted noise into the data. It is assumed that an adversary's knowledge may encompass all, or part, of the following: details of the learning algorithm; details of the parameters (e.g., feature weights); and feedback on decisions. Analysis assuming perfect knowledge provides an upper bound on the performance degradation that can occur under attack. It is however usually fair to assume that the adversary is not totally unconstrained---in particular, it is a common assumption that an attacker can only control a given small fraction of the training samples; and in the case of test-time attacks can only enforce up to a certain maximum number of modifications. 

Based on the adversary's knowledge about the targeted ML model, we can classify adversarial attacks into two categories: (1) a \textit{white-box attack}, and (2) a \textit{black-box attack}. In white-box attacks, it is assumed that the adversary has complete knowledge of the ML model including its architecture, training data, and the hyper-parameters. In black-box attacks, the ML model is an opaque black box for the adversary and no information (in terms of the technique being used or the hyper-parameters) is assumed to be available. It is however granted in black-box attacks that the adversary can operate as a standard user who can query the system with examples and note the model response. These query/response pairs can then be used by the attacker to infer the ``ML black-box'' and adversarial examples can be accordingly crafted. 


Most of these adversarial ML attacks operate according to the following two-step methodology. Firstly, an attacker chooses the target image/images or target output class/classes (in the case of targeted misclassification) and defines the optimization goals, i.e., correlation coefficients, accuracy or other parameters to analyze imperceptibility. Secondly, a random noise is introduced into the target image to compute the imperceptibility based on the defined optimization goals. If the optimum imperceptibility is achieved, the intruded image is considered as an adversarial image; otherwise, the noise is updated based on imperceptibility parameters and a new image is generated. 

Many adversarial attacks have been proposed in the literature such as the limited-memory Broyden-Fletcher-Goldfarb-Shanno (L-BFGS) method \cite{szegedy2013intriguing,tabacof2016exploring}; the fast gradient sign method (FGSM) \cite{rozsa2016adversarial,goodfellow2015explaining,kurakin2016adversarial} method; the basic iterative method (BIM); the Jacobian-based Saliency Map Attack (JSMA) \cite{papernot2016limitations}; the one-pixel attack \cite{su2017one}; the DeepFool attack \cite{moosavi2016deepfool}; the Zeroth Order Optimization (ZOO) attack \cite{chen2017zoo}; the CPPN EA Fool \cite{nguyen2015deep}; and the C\&W's attack \cite{carlini2017towards}. However, to study the effects of pre-processing noise filters on adversarial examples, in this paper, we limit ourselves to the most commonly used attacks, i.e., the L-BFGS, FGSM, BIM attack methods, which we briefly discuss next.


\vspace{1mm}
\subsubsection{The L-BFGS Attack}

This method is proposed by Szegedy et al. to generate an adversarial example in DNNs \cite{szegedy2013intriguing,tabacof2016exploring}. The basic principle of the L-BFGS method is to achieve the optimization goal as defined in Equation \ref{Eq:eq1}, where noise represents the perturbations and minimizing it represents its imperception. The main \textit{limitation of the L-BFGS method} is that it utilizes a basic linear search algorithm to update the noise for optimization which increases its converging time.

\begin{equation}
\min ||noise||_2 \implies f(x+noise) \neq f(x)
\label{Eq:eq1}
\end{equation}


\vspace{1mm}
\subsubsection{The FGSM Attack} 
 To address the computational cost issue, Goodfellow et al. proposed the FGSM algorithm for generating adversarial examples with fewer computations by performing a one-step gradient update along the direction of the sign of gradient at each pixel \cite{rozsa2016adversarial,goodfellow2015explaining} Their proposed imperceptive noise can be defined as $\eta = \epsilon \bigtriangledown_x J (\theta, x, f)$, where, $\epsilon$ and $\eta$ are the magnitude of the perturbation and the imperceptible noise, respectively. $J$ is the cost minimizing function (based on original image $x$, classification function $f$ and cost with respect to target class $\theta$) obtained through stochastic gradient descent. So, the generated adversarial example can be computed by adding $\eta$ into the targeted image. The main \textit{limitation of the FGSM method} is that these attacks are robust for white box attacks rather than for the black box attack. Several variants of FGSM were proposed to handle the white box assumption but increases it converging time which limits its applicability in real-world applications \cite{tramer2017ensemble}.


\vspace{1mm}
\subsubsection{The BIM Attack}
Previous methods assume adversarial data can be directly fed into the DNNs. However, in many applications, people can only pass data through devices (e.g., cameras, sensors). Kurakin et al. applied adversarial examples to the physical world \cite{kurakin2016adversarial} by extending the FGSM algorithm by running a finer optimization (smaller change) for multiple iterations. In each iteration, the authors proposed to clip the pixel values to avoid large changes on each pixel. The main \textit{limitation of the BIM method} is that it ignores the pre-processing stages after acquiring the data.


\section{Effect of pre-processing noise filtering on Adversarial ML}
\label{fademl}

\begin{figure}[!t]
	\centering
	\includegraphics[width=.8\linewidth]{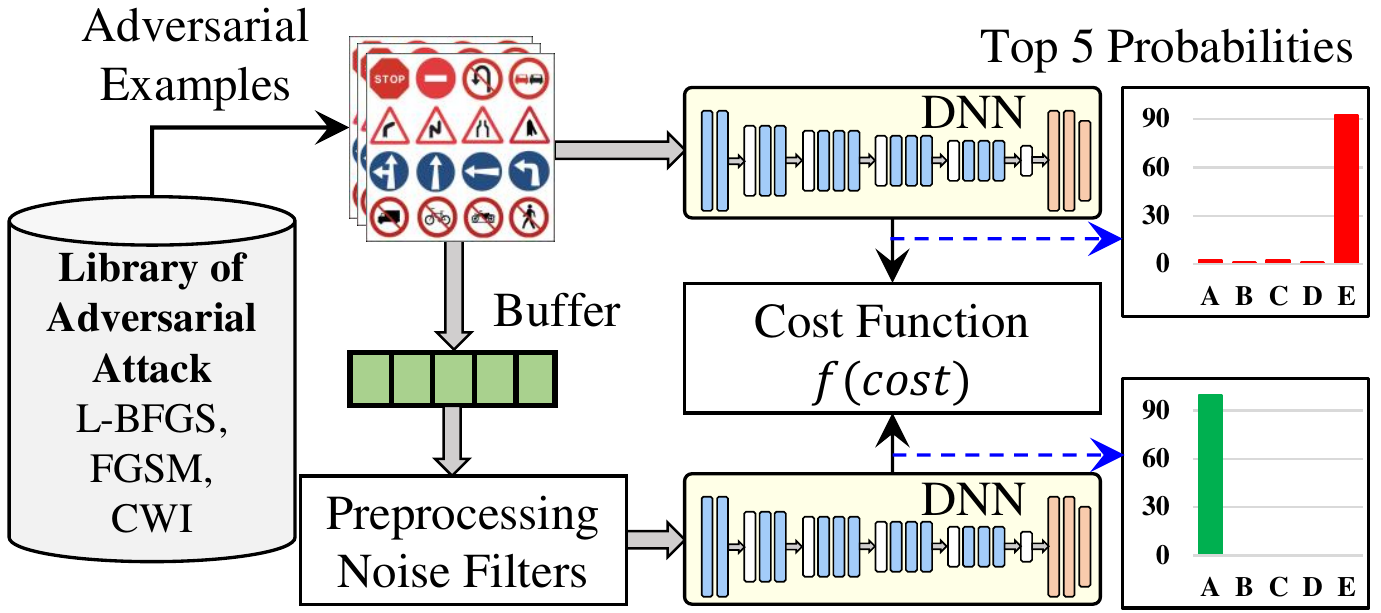}
	\caption{\textit{Proposed analysis methodology to analyze the impact of the pre-processing noise filtering on Adversarial ML methods.}}
	\label{fig:FAdeML_analysis}
\end{figure}

\begin{figure}[!t]
	\centering
	\includegraphics[width=.9\linewidth]{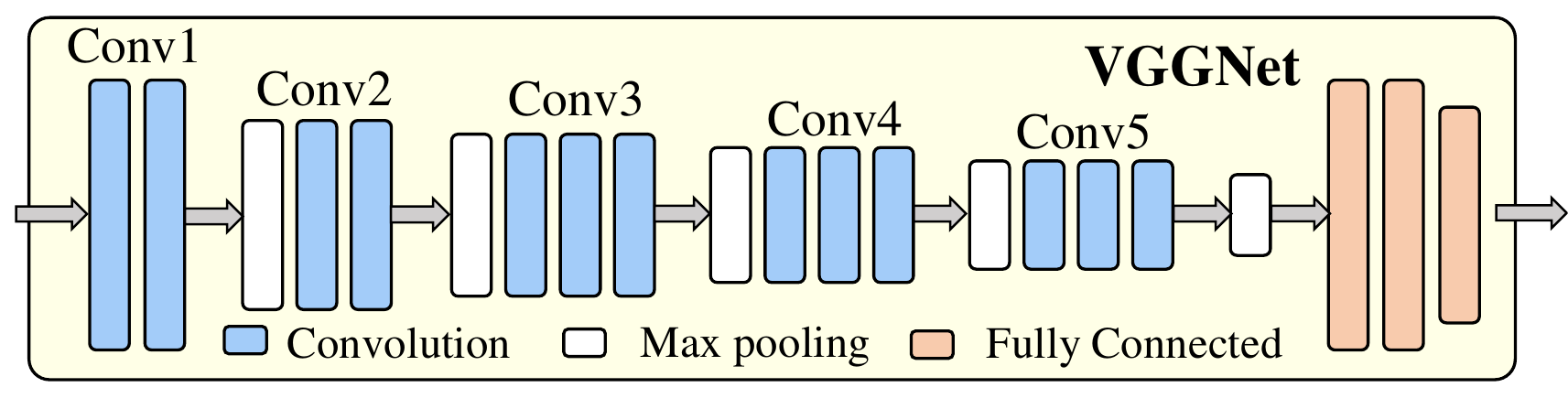}
	\caption{\textit{VGGNet: Conv1 (64 output filters), Conv2 (128 output filters), Conv3 (256 output filters), Conv4 (512 output filters) and Conv5 (512 output filters)}}
	\label{fig:vggnet}
\end{figure}

\begin{figure*}[!t]
	\centering
	\includegraphics[width=1\linewidth]{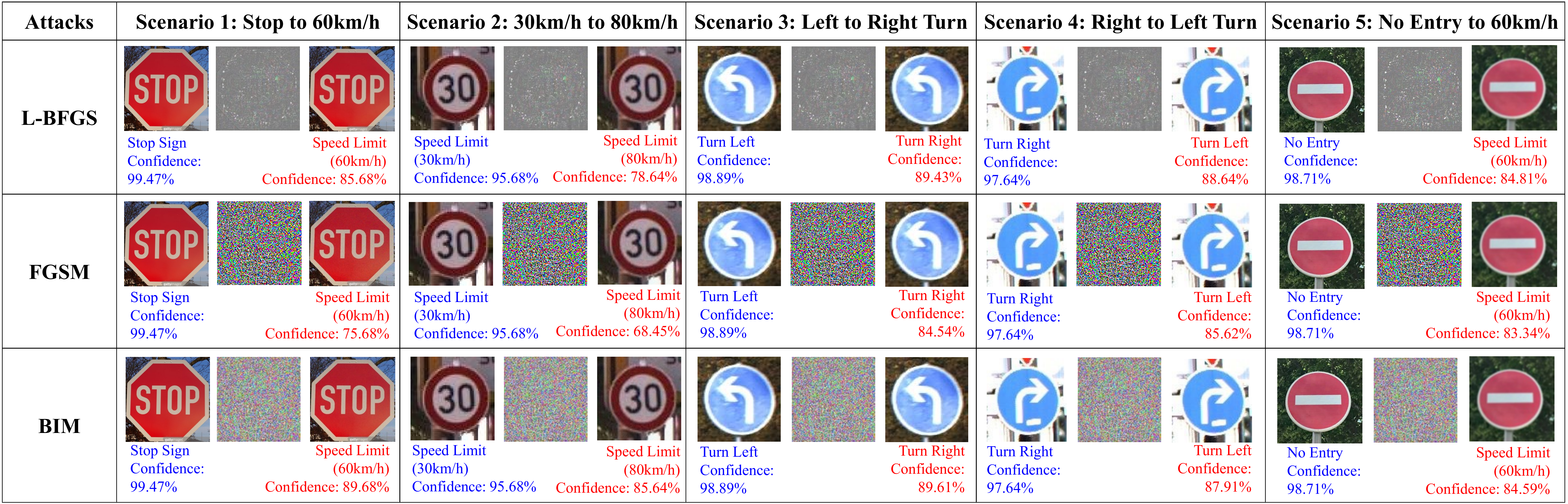}
	\caption{\textit{The impact of implemented adversarial attacks, i.e., L-BFG, FGSM and BIM, for misclassification \textbf{under the assumptions of Threat Models I,} on the top 5 accuracy of VGGNet trained on GTSRB.}}
	\label{fig:FAdeML_analysis_TMI}
\end{figure*}

In this section, we propose our analysis methodology for studying the effects of the pre-processing noise filter on adversarial security attacks during the ML inference. Our analysis methodology comprises the following steps (also illustrated in Fig. \ref{fig:FAdeML_analysis}):

\begin{enumerate}

    \item We initially choose an attack method from the adversarial attack library to generate adversarial examples according to their corresponding optimization functions and scaling methods.
    
    \item Inference is then performed on a trained DNN model to compute the classification probabilities for the generated adversarial examples \textit{assuming Threat Model I} (which assumes that the attacker can access the pre-processing noise filter's output). 
    
    \item Inference is also performed on the same trained DNN  model using the same adversarial examples, but now assuming \textit{Threat Model II or III}, and classification probabilities are computed.
 
    \item Finally, we compare the difference between the classification probabilities computed under the assumptions of Threat Models I and II/III, using the following cost function:

    \begin{equation}
        f(cost) = \sum_{n = 1}^{5} P (C_n) - P (C^*_n)
        \label{Eq:eq3}
    \end{equation}

    where, $C_n$ and $C^*_n\in \{C_1, C_2, ..., C_5\}$, $P(C_n)$ and $P(C^*_n)$ represent the top five predictions/classes for particular input/adversarial example, under the assumptions of the Threat Models I and II/III and their respective classification probabilities. 

\end{enumerate}

\subsection{Experimental Setup}\label{exp_setup}
To demonstrate the effectiveness of our analysis methodology, and to analyze how state-of-the-art adversarial attacks perform in the presence of noise filters, the following experimental setup is used:

\begin{enumerate}
    \item \textbf{DNN:} we use VGGNet model, which is composed of five convolutional layers and one fully connected layer (Fig. \ref{fig:vggnet}).
    
    \item \textbf{Pre-processing Noise Filter:} We implement the filters \textit{local average with neighborhood pixels} (LAP) and \textit{local average with radius} (LAR). For comprehensive analysis, we use five distinct configurations of LAP each for different number of neighboring pixels $np = {4, 8, 16, 32, 64}$ and radius $r = {1, 2, 3, 4, 5}$.  
    
    \item \textbf{Dataset:} we use the German Traffic Sign Recognition Benchmarks (GTSRB) dataset 
    
    \item \textbf{Adversarial Attacks and Threat Models:} we use the L-BFGS, FGSM, and BIM attacks and assume Threat Models I and II/III.
    
    \item \textbf{Payload:} We perform five targeted misclassification scenarios, i.e., (i) stop sign to 60km/h; (ii) 30km/h to 80km/h; (iii) left turn to right turn; (iv) right turn to left turn; and (v) no entry to 60km/h.
\end{enumerate}

\begin{figure}[!t]
	\centering
	\includegraphics[width=1\linewidth]{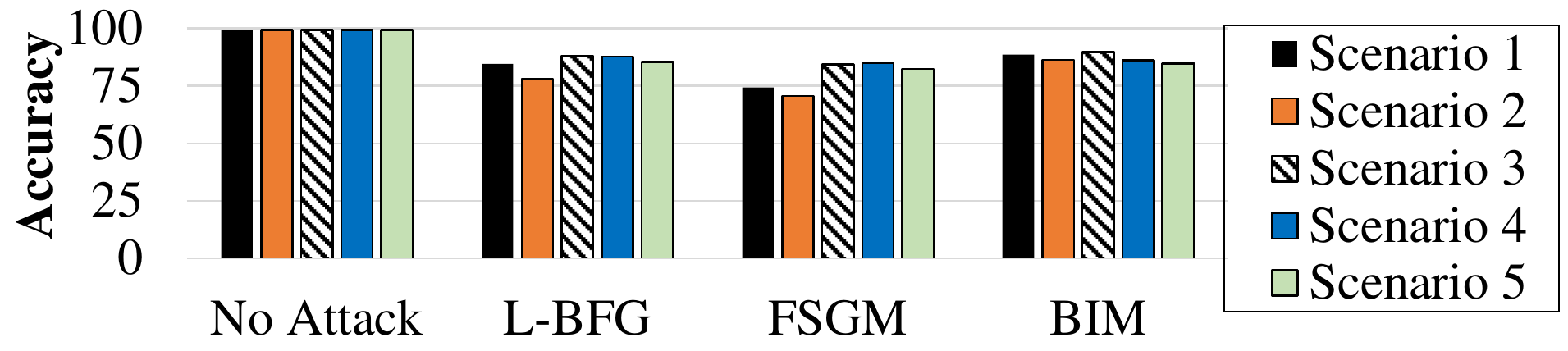}
	\caption{\textit{Top 5 accuracy of overall VGGNet, a DNN trained on GTRSB dataset, for different adversarial attacks, i.e., L-BFG, FGSM
	, BIM }}
	\label{fig:FAdeML_analysis_WF}
\end{figure}

\begin{figure*}[!t]
	\centering
	\includegraphics[width=1\linewidth]{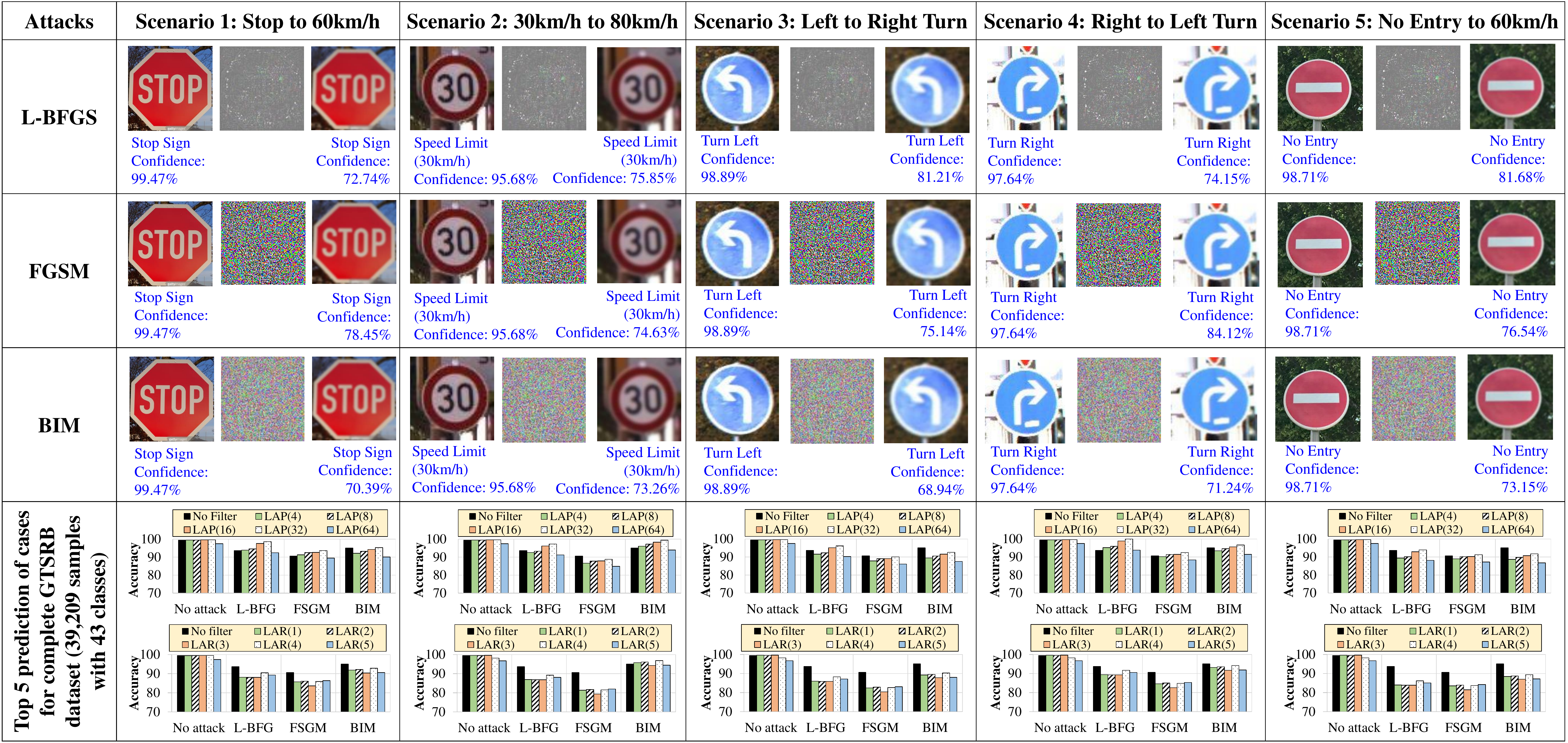}
	\caption{\textit{The traditional adversarial attacks, i.e., L-BFG, FGSM and BIM, are neutralized by pre-processing low pass filters, i.e., LAP and LAR, (\textbf{under the assumptions of Threat Models II and III}), in the expense of confidence reduction.}}
	\label{fig:FAdeML_analysis_TMII/III}
\end{figure*}

\subsection{Experimental Analysis}

In this section, we discuss the two analyses that we performed based on the aforementioned experimental setup under the assumptions of Threat Model I (Fig. \ref{fig:FAdeML_analysis_TMI}) and II/III (Fig. \ref{fig:FAdeML_analysis_TMII/III}), for the selected adversarial attacks. The analysis in Fig. \ref{fig:FAdeML_analysis_TMI}, \textit{which assumes the Threat Model I}, shows that the implemented adversarial attacks successfully performed misclassfication for all the attacking scenarios. Though, adversarial noise is invisible but still the adversarial examples have an (up to 10\%) on the overall top 5 accuracy of the VGGNets for complete GTSRB dataset. The analysis in Fig. \ref{fig:FAdeML_analysis_TMII/III}, \textit{which assumes the Threat Models II/III}, however shows that the smoothing filters (LAP and LAR) can nullify the impact of the adversarial examples on classification, but the attack is shown to affect the top 5 accuracy of overall DNN. The detailed discussion and insights of this is given below:   

\begin{enumerate}
    \item \textbf{L-BFGS and FGSM:} The LAP filters nullify the effects of L-BFGS filters but the top 5 accuracy is also reduced by 10\%, as shown in Fig. \ref{fig:FAdeML_analysis_TMI}. However, if we increase the number of ``$np$'' then the accuracy improves but it decreases after $np = 32$. In the case of LAR filters, the top 5 accuracy starts decreasing after $r = 4$. The impact of LAP and LAR filters on FGSM attacks follows similar trends. This impact is more pronounced on L-BFGS because its optimization function is highly dependent on the sharp edges or sudden changes in data samples, which are removed by smoothing filters. Moreover, both of these attacks do not consider the effect of pre-processing while designing adversarial examples. Therefore, in the presence of smoothing filters, they still reduce confidence of the network by decreasing the top 5 accuracy.
    
    \item \textbf{BIM:} This analysis shows the smoothing filters also nullify the effects of this attack. Since, this attack considers the feedback from the pre-processing, therefore, the effects of this attack is nullified when smoothing strength is relatively high. This has a significant impact on the confidence even after filtering effects, as shown by the confidence values of all scenario in Fig. \ref{fig:FAdeML_analysis_TMII/III}, similarly, the top 5 accuracy of the network increases with increase in number of neighboring pixel (till $np$ = 32) or radius ($r$ = 3) for local average filters. However, further increase in the neighboring pixel (till $np$ $>$ 32) or radius ($r$ $>$ 3) decreases the top 5 accuracy because it compromises some of the distinguishing features as well, as shown in the top 5 accuracy analysis of Fig. \ref{fig:FAdeML_analysis_TMII/III}.   
\end{enumerate} 

\subsection{Key Insights}

\begin{enumerate}
    
    \item The adversarial noise from the adversarial examples that are gradient descent based can be removed by applying the smoothing filters, i.e., LAR  or LAP. However, the attacks do still reduce the confidence of the overall classification. 
    
    \item The selection of smoothing parameters in LAR or LAP has a direct impact on the top 5 accuracy of the DNN. Top 5 accuracy increases with increase in smoothing parameters till a certain value, e.g., for LAP and LAR values are $np = 32$ and $r=4$. When going beyond this threshold value, the top 5 accuracy starts to decrease because of the degradation of input image/sample quality.
    
    \item To perform a successful adversarial attack, the effects of all the pre-processing stages should be considered while optimizing the adversarial noise.
\end{enumerate}
  
\section{Pre-processing Noise Filtering-aware Adversarial ML}
\label{FAde_attack}

The methodology of our proposed pre-processing noise filter aware FAdeML attack comprises the following steps, shown in Fig. \ref{fig:FAdeML_attack}:
\begin{figure}[!t]
	\centering
	\includegraphics[width=1\linewidth]{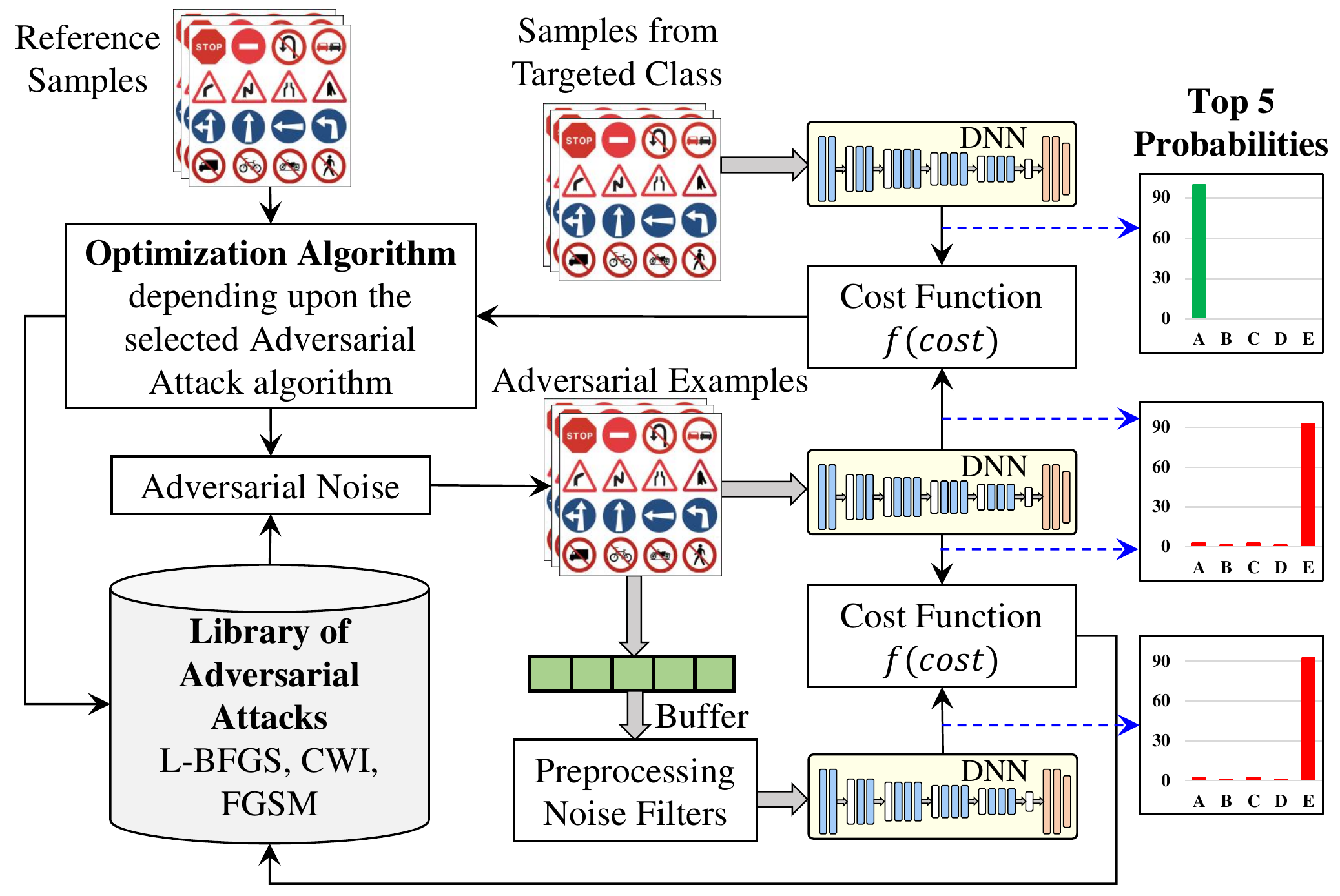}
	\caption{The proposed methodology to design FAdeML Attacks to incorporate the Impact of Pre-processing Noise Filtering on Adversarial ML}
	\label{fig:FAdeML_attack}
	\vspace{-0.3cm}
\end{figure}
\begin{figure*}[!t]
	\centering
	\includegraphics[width=1\linewidth]{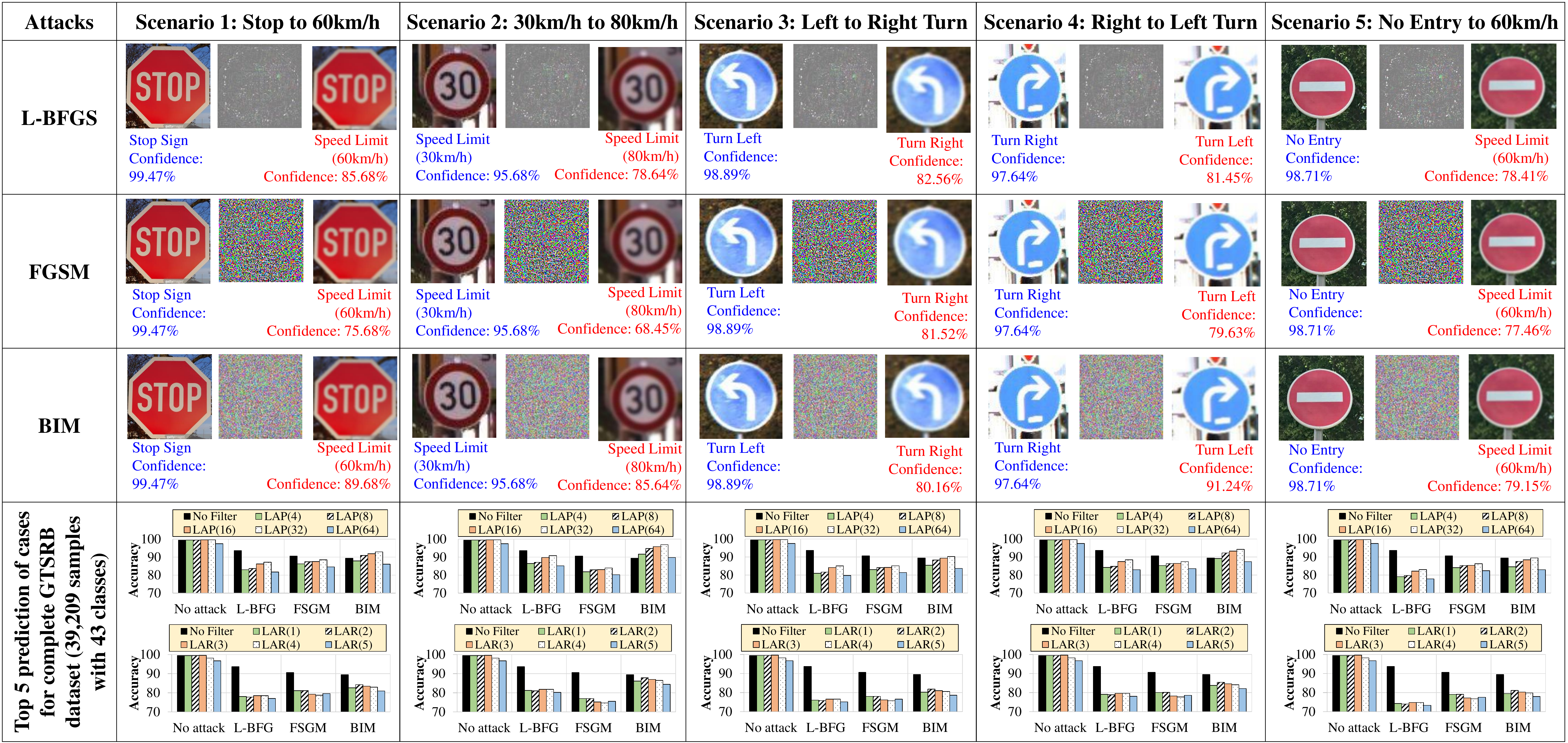}
	\caption{\textit{ Unlike the the traditional adversarial attacks, \textbf{FAdeML attacks} \textit{are not neutralized by pre-processing low pass filters} but their impact on top 5 accuracy of overall neural network is relatively higher.}}
	\label{fig:FAdeML_attack_TMI}
\end{figure*}
\begin{enumerate}
    
    \item We initially choose a reference sample $``x"$ to be perturbed and a sample of the targeted class $``y"$ for the misclassification such that $prediction (x) \neq prediction (y)$, and choose an attack from the adversarial attack library.
    
    \item We compute the prediction/classification probabilities for sample $``x"$ and $``y"$, \textit{assuming Threat Model I}, to identify the difference between their respective prediction/classification probabilities by computing the following cost function: $f(cost) = \sum_{n = 1}^{5} P_x (C_n) - P_y (C^*_n)$
    
    
    \item We compute and add the adversarial noise $``n"$ to $``x"$ to generate the adversarial example $``x^*"$ (i.e., $x^* = \eta \times n + x$), where, $\eta$ is the noise scaling factor to make it imperceptible. 


    \item We then compute the prediction/classification probabilities for $``x^*"$, \textit{assuming Threat Model II or III}, to identify the impact of pre-processing noise filtering on generated adversarial example $``x^*"$.
    
    \item We then analyze the difference between the prediction/classification probabilities of adversarial example $``x^*"$, assuming Threat Models I and II/III, based on the cost function of Equation \ref{Eq:eq3}.
    
    \item Finally, we repeat the optimization depending upon the selected algorithm to incorporate the impact of pre-processing noise filtering.
   
    \begin{equation}
        x^* = \eta \times (n + \frac{\delta n}{\delta f(cost)}) + x 
        \label{Eq:eq6}
    \end{equation}

\end{enumerate}

\subsection{Experimental Analysis}
To illustrate the effectiveness of the proposed FAdeML attack methodology, we utilize the same experimental setup (discussed in Section \ref{exp_setup}). We performed experiments to those shown in Fig. \ref{fig:FAdeML_attack_TMI} for the proposed FAdeML attacks, and identified the following key observations: 

\begin{enumerate}
    
    \item FAdeML attack performs better even in the presence of pre-processing filters, because these attacks incorporates the smoothing effects of low pass pre-processing noise filters (i.e., LAP or LAR). 
    
    \item Another key observation is that the attack confidence reduces slightly because the smoothing effect also compromises the sudden changes, which are then exploited by most of the adversarial attacks. 
    
    \item Similar observation can be depicted from the analysis presented in Fig. \ref{fig:FAdeML_attack_TMI}. It shows that top 5 accuracy of the network increases with increase in number of neighboring pixel (till np $=$ 32) or radius (r = 3) for local average filters. However,further increase neighboring pixel (till np $>$ 32) or radius (r $>$ 3) decreases the top 5 accuracy because it compromises some of the distinguishing features as well.  
\end{enumerate} 

\subsection{Key Insights}

\begin{enumerate}
    \item The effectiveness of the adversarial noise can be increased significantly by optimizing it with respect to pre-processing stages, especially noise filtering.  
   
    \item The effects of gradient-descent function optimized adversarial noise can be nullified because they rely heavily on the sudden changes in the input images/samples.
\end{enumerate}

\section{Conclusion}\label{conclusion}

In this paper, we proposed an analysis methodology to understand the impact of pre-processing filter on adversarial attacks and an attack methodology to incorporates these effects into the adversarial examples. We have demonstrated the effectiveness of the proposed analysis and attack methodology using three state-of-the-art attacks (L-BFG, FGSM and BIM) on the ``VGGNet'' Deep Neural Network (DNN) trained on the ``German Traffic Sign Recognition Benchmarks (GTSRB)'' dataset. The experimental results shows that the effects of existing adversarial attacks on classification can be nullified by applying the smoothing filters, LAP and LAR, on the input samples before sending it the DNNs, even though the confidence classification is still affected. Based on this analysis, we developed a new pre-processing noise-\textit{F}ilter-aware \textit{Ad}v\textit{e}rsarial ML (FAdeML) attack and showed that FAdeML is able to force a misclassification even after the application of the smoothing filters. Although previous work in the related literature has explored using pre-processing for defense purposes, we are the first to explicitly exploit noise filtering to improve an adversarial ML attack. We will open-source our designs. Our overall goal through this work is to communicate to the community the potential vulnerabilities of current ML systems and thereby inspire researchers to develop ML architectures that are effective yet can resist adversarial examples.



\bibliographystyle{IEEEtran}
\bibliography{bib/bibliography}

\end{document}